%% file: main.tex
\documentclass{article} % For LaTeX2e
\usepackage[a4paper, margin=2cm]{geometry}
\usepackage[T1]{fontenc}
\input{math_commands.tex}

\RequirePackage{natbib}

\setcitestyle{authoryear,round,citesep={;},aysep={,},yysep={;}}

\usepackage{hyperref}
\usepackage{url}
\usepackage{todonotes}
\usepackage{amsmath}
\usepackage{enumitem}
\usepackage{amsthm}
\usepackage{comment}
\usepackage{caption}
\usepackage{subcaption}

\usepackage[autostyle=try, english=american]{csquotes}
\MakeOuterQuote{"}

\newtheorem{proposition}{Proposition}

\title{Resolving Spurious Correlations in\\ Causal Models of Environments via Interventions}

\author{Sergei Volodin, Nevan Wichers, Jeremy Nixon \\
Google Research\\
\texttt{\{volodin,wichersn,jeremynixon\}@google.com} \\
}

\newtheorem{definition}{Definition}

\begin{document}

\maketitle

\begin{abstract}
Causal models bring many benefits to decision-making systems (or agents) by making them interpretable, sample-efficient and robust to changes in the input distribution. However, spurious correlations can lead to wrong causal models and predictions.
We consider the problem of inferring a causal model of a reinforcement learning environment and we propose a method to deal with spurious correlations.
Specifically, our method designs a reward function which incentivises an agent to do an intervention to find errors in the causal model. The data obtained from doing the intervention is used to improve the causal model.
We propose several intervention design methods and compare them.
The experimental results in a grid-world environment show that our approach leads to better causal models compared to baselines: learning the model on data from a random policy or a policy trained on the environment's reward.
The main contribution consists of methods to design interventions to resolve spurious correlations.
\end{abstract}

\section{Introduction}
Reinforcement Learning \citep{sutton2018reinforcement} (RL) is a general framework where an agent interacts with an environment to optimize a reward. The recent successes \citep{mnih2013playing,silver2016mastering,berner2019dota,vinyals2019grandmaster,akkaya2019solving} of applying RL algorithms to games sparked an interest in the framework as a general solution to Artificial Intelligence problems. However, current RL agents have significant shortcomings \citep{rlblogpost,kurenkov2018reinforcementflaw}, such as lack of interpretability, lack of robustness to distributional shift, susceptibility to adversarial examples \citep{gleave2019adversarial}, and high sample complexity.

One of the frameworks that can arguably solve these issues is Causal Reasoning \citep{pearl2018book,halpern2005causes,pearl2018theoretical}. The framework considers a system whose states change with time. The goal is to model this system as a set of equations which predict the next state given the history.

One of the problems that causal reasoning considers is the inference of the model given limited data from a dynamical system \citep{javed2020learning}. When solving this problem, spurious correlations are a major issue. They are pairs of variables which are associated given the data but not actually causally related. For example, we might care about predicting when the Sun rises. If we hear a rooster making a sound every time before the Sun rises, we might learn to rely on that feature. However, this would be a spurious correlation, since the sound does not actually {\em cause} the Sun to rise. If we move to an area where we cannot hear the rooster, we will know that it is the case: the Sun rises despite the fact that we do not hear the rooster.

To resolve this issue, we need to obtain a more diverse dataset to train the causal model. To do so, we need to perform an {\em intervention} \citep{halpern2005causes} in the environment, such as moving to another location. In general, interventions are sets of actions that change the values of some variables in the causal model. They are used to create a better causal model. Data from an intervention can cause a spurious correlation to disappear from the dataset in a sense that it is no longer beneficial to rely on the spurious correlation. Therefore, a model trained on the new dataset would be forced to use more reliable features, for example, the current relative position of the Sun and the Earth.

{\bf Contribution.} The main contribution consists of methods to design interventions to resolve spurious correlations in Causal Models of Reinforcement Learning environments (Section \ref{sec:solution}). In addition, we formulate the problem of learning a causal model of an RL environment mathematically in Section \ref{sec:problem}. The metric of success is how well we can uncover the correct causal model, rather than the final performance of the agent.

{\bf Method overview and steps.} In this paper, we consider the causal model of the environment that the agent is deployed into. To do so, we train the model on the rollouts data obtained from agent-environment interaction. Our approach collects rollouts, then trains a causal model on that data, then computes the (approximate) best reward function to uncover spurious correlations. This reward depends on the causal model and encourages the agent to perform structured interventions in the environment. We thus call this reward an {\em intervention reward}. We propose several methods calculate the intervention reward: from hand-designed heuristics to end-to-end learned methods. After training a policy for the intervention reward, we collect data from rollouts of the agent's policy. Then we train the causal model on this new data to learn a model with fewer spurious correlations. Then a new intervention is designed based on the new causal model to repeat the cycle.

We would like to understand the relationship between high-level concepts in the environment, such as positions of players, rather than low-level pixel representation. So, our causal model uses {\em computed features} rather than raw observations. In our setup, the features are computed from observation by a hand-designed function.

\begin{figure}[t]
    \centering
        \begin{subfigure}[b]{0.78\textwidth}
         \centering
         \includegraphics[width=\textwidth]{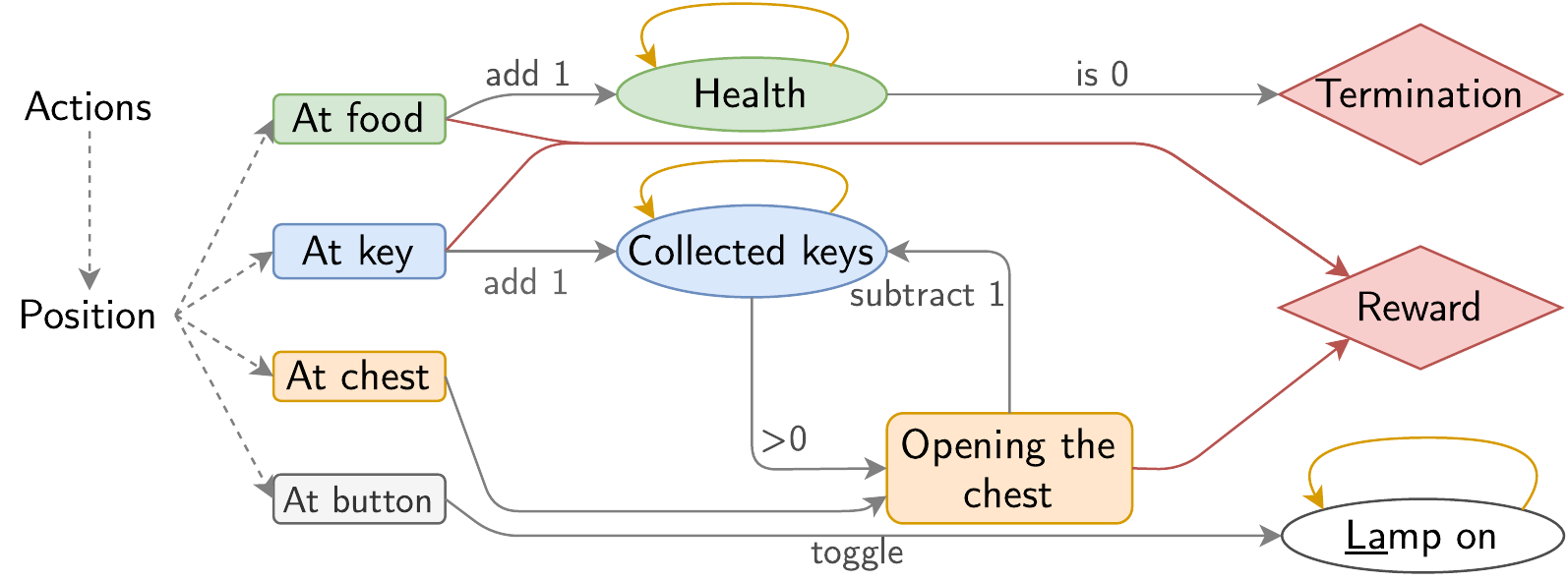}
         \caption{Causal diagram for the KeyChest environment}
         \label{fig:envs_causal}
    \end{subfigure}
    \hfill
    \begin{subfigure}[b]{0.09\textwidth}
         \centering
         \includegraphics[width=\textwidth]{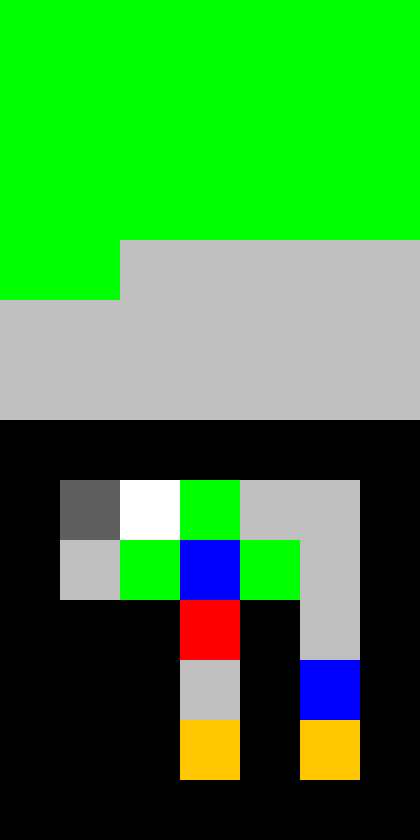}
         \caption{Env. B}
         \label{fig:envs_b}
        %  \hspace{50pt}
        \vspace{20pt}
    \end{subfigure}
    \hfill
    \begin{subfigure}[b]{0.11\textwidth}
         \centering
         \includegraphics[width=\textwidth]{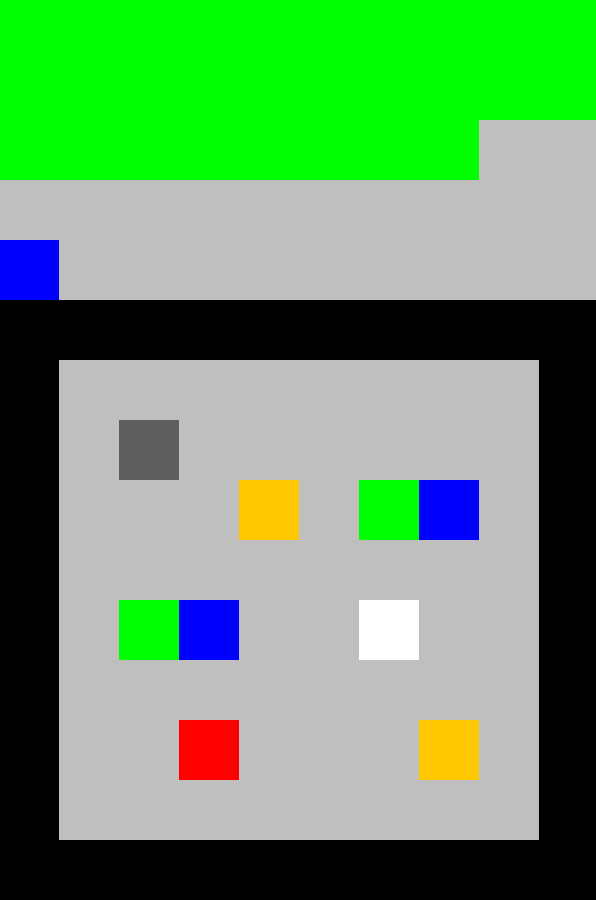}
         \caption{Env. C}
         \label{fig:envs_c}
         \vspace{20pt}
    \end{subfigure}
    
    \caption{Left \ref{fig:envs_causal}: The causal diagram of the environment which the agent should learn.  {\color{red} The player} needs to collect {\color{green} food} and {\color{blue} keys}. Keys are used to open {\color{orange} chests} and the {\color{blue} number of keys} is displayed above the first black line on Figure \ref{fig:envs_c} (and it is 0 in Figure \ref{fig:envs_b}). {\color{green} Top row with health} decreases at every time-step, and the episode ends if it is 0. {\color{gray} The button} toggles the lamp (\underline{black}/white) which gives no reward. {\color{orange} Orange} arrows going to the left show that the variable depends on itself at the previous time-step. {\color{red} Red} arrows indicate that the node results in a reward. All other arrows mean that the child depends on the value of the parent at the previous time-step, except for the node Opening the Chest, which depends on values at the current time-step. The right two figures \ref{fig:envs_b}, \ref{fig:envs_c} show the layouts of environments B and C. The agent is {\color{red} red}, food is {\color{green} green}, keys are {\color{blue} blue}, chests are {\color{orange} orange}, button is {\color{gray} gray}, and the lamp is either white or black (white on the figures). The dynamics of the environment is described in Section \ref{sec:environment}.}
    \label{fig:envs}
\end{figure}

\section{Problem}
\label{sec:problem}
We focus on the problem of inferring a causal graph (causal model) of the environment in a reinforcement learning setting.

\subsection{Notation}
We use $[n]$ to denote the set of first $n$ positive integers: $[n]=\{1,2,3,...,n\}$. $x\sim X$ means sampling a random variable $x$ from a distribution $X$.

{\bf Reinforcement learning.} We consider the Reinforcement Learning setup consisting of two entities: the agent with a policy $\pi$ and the environment $\mu$. They interact with each other: the environment gives the agent an observation $o\in\mathcal O$, then the agent gives an action $a\in\mathcal A$ and the environment responds with a reward $r\in\mathcal R$. We call $x=(o,a,r)$ a step. After one step, the cycle is repeated. The environment begins and ends the interaction. We call one interaction an episode (or a {\em rollout}) with {\em history} $h=((o_1,a_1,r_1),\ldots,(o_T,a_T,r_T))$ where $T$ is the {\em length} of the episode. Sometimes we call multiple episodes $(h_1,...,h_k)$ a {\em history} as well. A {\em policy} $\pi$ is a mapping from observations to a distribution over actions. By $(\mu,\pi)$ we denote the distribution of histories obtained from interactions between the environment $\mu$ and the policy $\pi$. 

{\bf Causal models.} We take the standard definition of Functional Causal models from \cite{pearl2009causality} (page 26). We consider a set of variables $f_i$ which are evolving in time. The model is a directed graph $G$ with variables as nodes. The value at the next time-step $t+1$ for a variable $f_i$ is determined by a function of its parents ${\bf PA}(f_i)$ in the graph: $f_i^{t+1}=M_i(f_j^{t},\,j\in {\bf PA}(f_i))$. Here $M_i$ is a fixed but unknown function.

{\bf Causal learning from step features.}
We want to do causal learning on higher level features \citep{nachum2018data} instead of the raw observations from the environment. Therefore, we extract high-level features from the observations using a hand designed function $f\colon\mathcal O\to\mathcal F$ with $\mathcal F=\mathbb R^F$, and do the causal learning on those extracted features. From this point on, we use features/nodes/variables $f_i$ interchangeably since they mean the same thing. Features for our setup are shown in Figure \ref{fig:envs_causal}.

We run a RL agent following a policy $\pi$ in an environment, which generates a history $h$ of observations $\mathcal O$. Then we compute features from those states using a hand-designed function $f$. Then, we learn a causal graph from that history by fitting the functions $M_i$ for each node to predict features at the next time-step given the current features. In our setup, functions $M_i$ are linear.  In addition, to reduce the number of edges in the model, we regularize $M_i$ for sparsity with an $l_1$ loss\footnote{Sparse causal models of dynamical systems for interpretability can be seen in \citep{bengio2017consciousness}}. This method is called {\em Granger} causality \citep{granger1969investigating}.

To learn the model, we consider all of the nodes as potential parents for any given node. After learning the model, we threshold the weights based on their magnitude to get the parents ${\bf PA}(f_i)$ for each node: edges in the graph correspond to weights of high magnitude, and non-existent edges correspond to weights close to $0$.

Mean squared loss of the causal model $G$ on histories data generated by interaction $(\mu,\pi)$ between an environment $\mu$ and an agent running a policy $\pi$ is defined as:
\begin{equation}
L_{\pi}(G)=\mathbb E_{h\sim (\mu,\pi)}\sum\limits_{t=1}^T\sum\limits_{i=1}^F\left(f_i^t-\hat{f}_i^t\right)^2
\label{eq:loss_def}
\end{equation}

Here $\tilde{f}_i^t=M_i(f_j^{t-1},\,j\in [n])$ is the predicted value for feature $i$ at time-step $t$ by the causal model $G$ in case if $t>0$ or a constant in case of $t=0$, the index $t$ represents the episode step number, and the random variable $T$ is the number of steps in the episode.

This loss measures how well the model can predict the features of the next time-step $f_t$ given previous features $f_{t-1}$. This is simply the loss of linear regression when fitting functions $M_i$ on the histories data. Note that $L=0$ corresponds to the perfect fit between the model and the true environment dynamics.

For a finite set of histories $\{h_e\}_{e=1}^E$ consisting of $E$ episodes sampled from $(\mu, \pi)$, we define the loss of the causal model $G$ as:
\begin{equation}
\label{eq:loss_def_finite}
L_{\pi,E}(G)=\frac{1}{E}\sum_{e=1}^E\sum_{t=1}^{T_e}\sum_{i=1}^F \left(f_i^{e,t}-\tilde{f}_i^{e,t}\right)^2
\end{equation}

Here $e$ is the episode number, $t$ is the step index in the episode, and $T_e$ is the number of steps in the episode with index $e$. Index $i$ indicates the feature, and $F$ is the number of features. $f_i^{e,t}$ thus is the feature $i$ at the time-step $t$ of episode $e$.

By the Central Limit Theorem, $L_{\pi}(G)=\lim_{E\to\infty}L_{\pi,E}(G)$.

\begin{figure}[t]
    \centering
    \includegraphics[width=\textwidth]{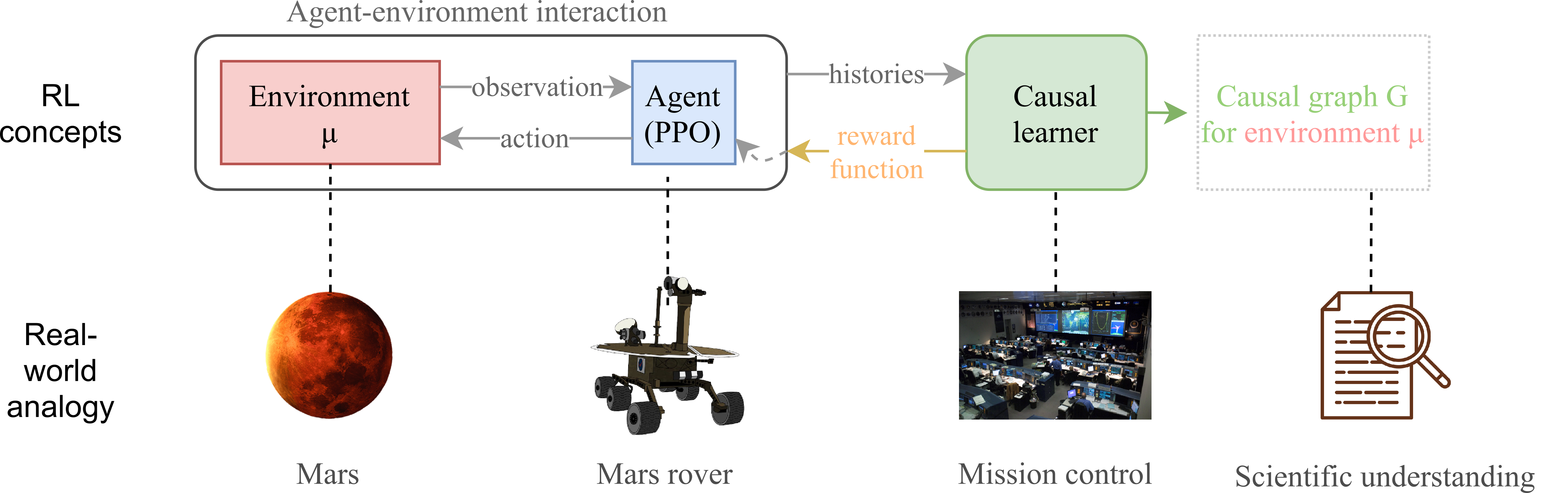}
    \caption{Problem setup. RL concepts (top row) are illustrated with real-world analogy (bottom row). Specifically, an environment $\mu$ is illustrated by a novel planet to explore, the agent corresponds to a rover on that planet executing low-level commands from the mission control, which corresponds to the causal learner that we introduce. At the end, the learner outputs a graph $G$ which corresponds to high-level understanding of the planet or the environment $\mu$}
    \label{fig:problem_setup}
\end{figure}

{\em Spurious correlations and interventions}. To sum up, we would like to design an algorithm to find a graph $G$ which fits the environment $\mu$, without giving the algorithm the ground truth graph $G^*$ but only the ability to interact with the environment. To do so, we can minimize the loss $L_{\pi, E}(G)$ over $G$ given histories generated by some policy $\pi$. For that, we run a policy $\pi$ in the environment to collect history $h$ and then we minimize $L$ on the histories.

Assuming we had a learning algorithm that could converge to $L_{\pi}(G)$ for any $\pi$, our Proposition \ref{prop:noiseless} (in the Appendix) shows that a random (uniform over actions) policy $\pi_r$ is sufficient to learn the graph in a realizable case defined by \citet{shalev2014understanding} where some graph $G^*$ perfectly fits the data\footnote{Note that in practice, a random policy might take too much time to explore the environment}.
This happens because, first, $G^*$ perfectly fits data from the random policy as well, so the graph $G^r=\arg\min_GL_{\pi_r}(G)$ found from $\pi_r$ would have zero loss on $\pi_r$ as well. Secondly, data from a random policy contains data from any policy, with some probability, therefore, $G^r$ has to fit data from all policies. In that case, interventions are not required to resolve spurious correlations: running a random policy is sufficient to uncover the causal structure of the environment.
One of the cases which falls into this category are fully-observable deterministic environments with one-hot encoding for states as features.

In contrast, in cases where we cannot fit the data perfectly ($L_{\pi,E}(G^*)>0$), it is possible that two policies produce different graphs, no matter the length of the history $t$ or the method to learn the graph.

\begin{figure}[t]
    \centering
    \includegraphics[width=0.9\textwidth]{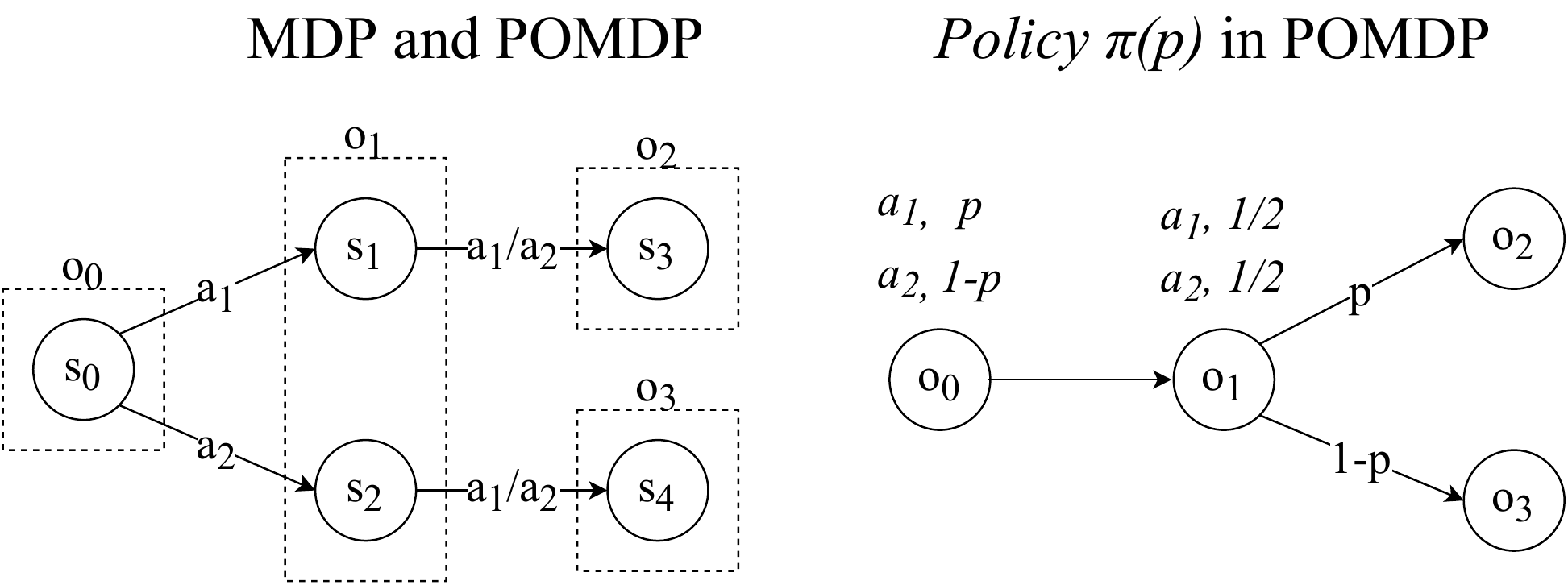}
    \caption{Environment with no ground truth graph $G$. In this example, the optimal graph depends on the policy collecting data. Specifically, given the observation $o_1$, the agent's prediction on what happens next depends on the action taken in $o_0$, because the true state (either $s_1$ or $s_2$) is inaccessible. The causal graph in this case depends on the policy's actions in state $o_0$ because by our assumption, we only use single time-step dependencies. {\bf Full description:} On the left, the state MDP (Markov Decision Process) is shown together with groupings of states resulting in equal observations. On the right, the resulting POMDP (Partially-Observable Markov Decision Process) after the grouping is shown, along with an example policy with parameter $p$, $\pi(p)$. In the MDP, there are 2 actions and 5 states: $s_0,\,s_1,\,s_2,\,s_3,\,s_4$. In $s_0$ (initial state) if we take $a_1$, we go to $s_1$. If we take $a_2$, we go to $s_2$. From $s_1$ and $s_2$, it does not matter which action is taken, as there is only one transition (to $s_3$ and $s_4$ respectively). $s_3$ and $s_4$ are terminal states. These states are grouped into 4 observations to form the POMDP shown on the right: $o_0=\{s_0\},\,o_1=\{s_1,\,s_2\},o_2=\{s_3\},\,o_3=\{s_4\}$. This means that if the agent receives an observation $o_1$, the environment is either in $s_1$ or in $s_2$, and the agent has no way to determine this. The policy $\pi(p)$ shown on the right takes action $a_1$ with probability $p$ in $o_0$, and takes $a_2$ with $1-p$ in $o_0$, and takes random uniform action in $o_1$. If we execute the policy $\pi(p)$ then the probability of going to $o_2$ or to $o_3$ from $o_1$ depends on this initial choice: we go to $o_2$ with probability $p$ and to $o_3$ with probability $1-p$. For features, we use 1-hot encoding for the states and actions. In Appendix \ref{app:pomdp1} we show that the optimal graph for policy $\pi(p)$ assigns probabilities $p,1-p$ respectively to future observations $o_2$, $o_3$ at state $o_1$.}
    \label{fig:ex_nonrealizable}
\end{figure}

Indeed, consider the MDP shown in Figure \ref{fig:ex_nonrealizable}. The causal graph $G^p$ minimizing $L_{\pi(p),t}(G^p)$ depends on the policy parameter $p$. In Appendix \ref{app:pomdp1} we show that for $p=0$ the optimal prediction for observation $o_1$ is to always predict $o_3$, and for $p=1$ the answer should be $o_2$. For $p\in(0,1)$ the optimal prediction should assign the values of $p,1-p$ for 1-hot encodings of $o_2$, $o_3$ respectively. This shows that the graph depends on the executed policy.

\begin{figure}[tbh]
    \centering
    \includegraphics[width=0.9\textwidth]{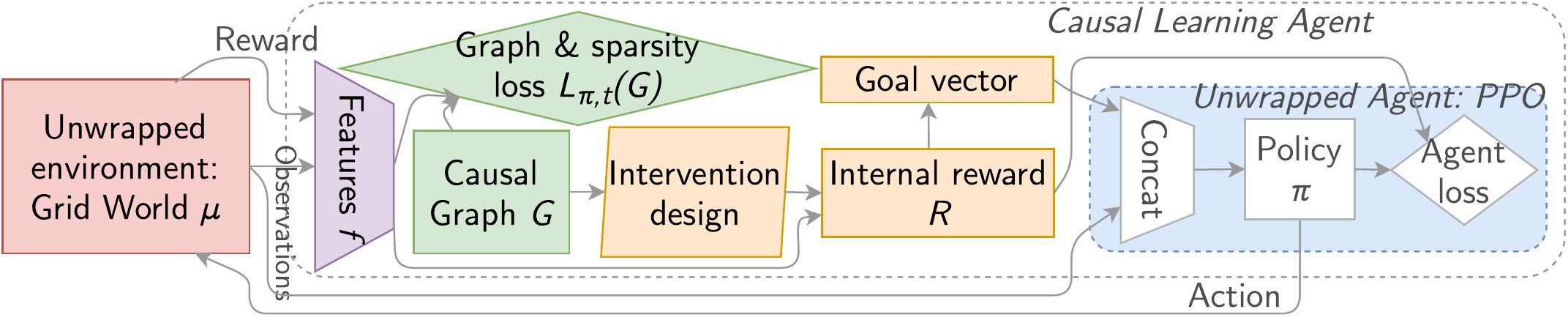}
    \caption{Learning the causal graph by actively interacting with the environment. Given a high-level set of {\color{violet} features $f_i$} and an {\color{red} environment $\mu$}, we collect data using a policy $\pi$ to learn an initial {\color{green} causal model $G$} from features time series. Then, we {\color{orange} design an {\em intervention reward}} which incentivises the agent to execute a policy which performs an intervention. The agent is trained on the intervention reward with standard {\color{blue} reinforcement learning.} The causal model is trained on the new interaction histories from the trained agent.}
    \label{fig:causalagent}
\end{figure}

\begin{figure}
    \centering
    \includegraphics[width=0.9\textwidth]{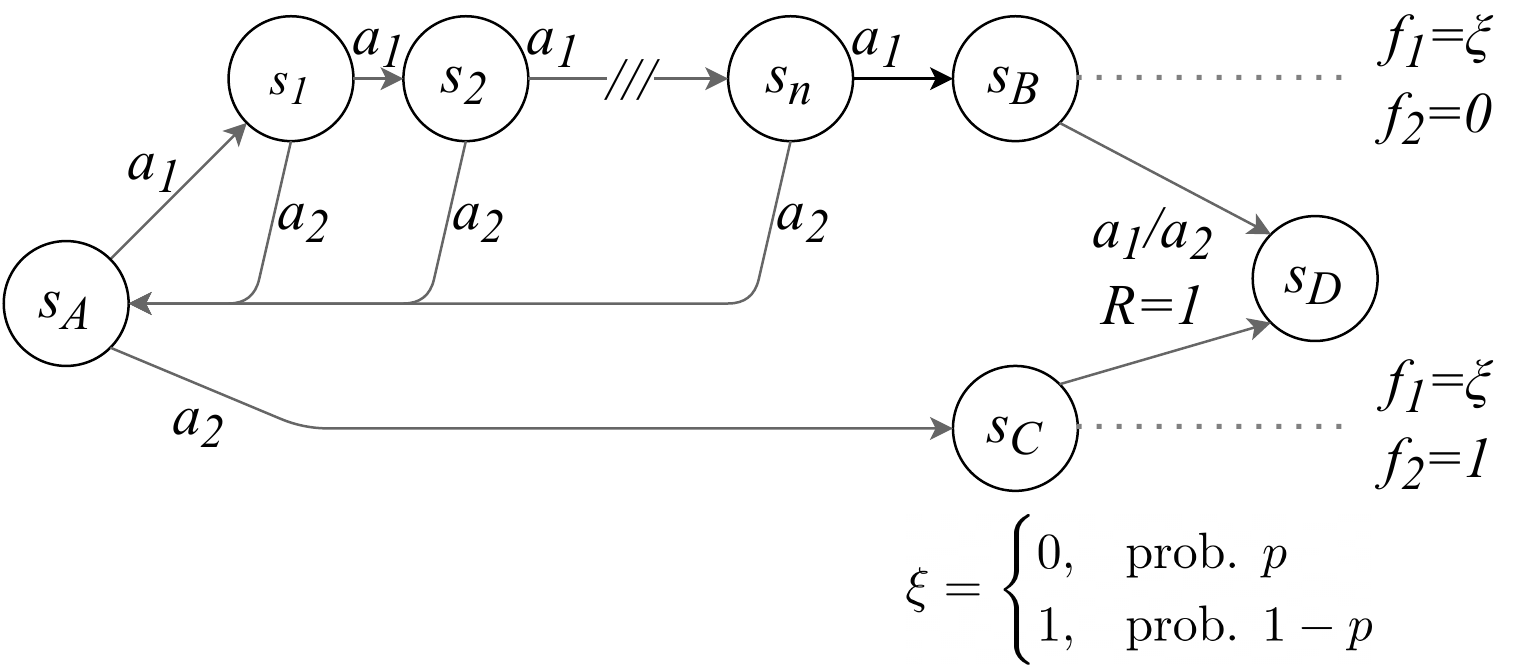}
    \caption{Environment where spurious correlations arise on a random policy but disappear in the minimax case. Specifically, we want to predict the reward $R$ from features $f_1,f_2$. These features are $0$ except for states $s_B$ and $s_C$. A large number $n$ leads to the random policy failing to ever visit $s_n$, which means that the causal graph will have much more data from $s_C$ than from $s_B$. Therefore, it will learn to use $f_2$ to predict the reward. In contrast, if we discover that we can go to $s_B$ by always executing action $a_1$, then the causal graph will favor relying on $f_1$, because doing so leads to a lower loss in both $s_B$ and $s_C$ when predicting the reward. Therefore, the minimax definition would rely on $f_1$. {\bf Full description:} In our MDP, there are $n+4$ states: $s_A,s_B,s_C,s_D,s_1,...,s_n$ with $s_A$ being the initial state. There are two actions, $a_1$ and $a_2$. The states are visible to the policy, but only the values of $f_1$, $f_2$ are visible to the causal learner. From the initial state, we go up if we take $a_1$ and continue going to the right if we take $a_1$ until we reach $s_B$. If we take $a_2$ in any of $s_1,...,s_n$, we return to $s_A$. If we take $a_2$ in $s_A$, we go to $s_C$. The reward is always $0$ except for the transitions $s_B\to s_D$ and $s_C\to s_D$ which both generate a reward of $1$ regardless of the action taken. The state $s_D$ is terminal. There are two additional features $f_1$ and $f_2$, and they are always zero except for $s_B$ where they take values $f_1=\xi$ and $f_2=0$, and $s_C$ where they are $f_1=\xi$ and $f_2=1$. The random variable $\xi$ is independent from all the other random variables and equals $1$ with probability $1-p$ and $0$ with probability $p$. We consider a policy which eventually arrives at $s_B$ with probability $q$ and at $s_C$ with probability $1-q$. The probability $q$ can take the value of $0$ if we take an action $a_2$ in $s_A$ or $1$ if we always take $a_1$. If we take random actions in $s_1,...,s_n$ and take $a_1$ in $s_A$, then this probability equals $\Theta (1/2^n)$. The fully random-uniform policy has $q=\Theta (1/2^n)$ as well.}
    \label{fig:ex_two_path}
\end{figure}

Now, to understand spurious correlations, consider the MDP on Figure \ref{fig:ex_two_path}. Consider the task of predicting the reward $R$ only from features $f_1, f_2$. If we rely on $f_1$, our model works regardless of being in $s_B$ or $s_C$ but there is a level of noise: the loss $L=p$. If we rely on $f_2$, our model works perfectly in $s_C$ but fails in $s_B$. In Appendix \ref{app:pomdp1} we show that a random uniform policy will result in relying on $f_1$ only for small enough $n \ll \log 1/p$. A random policy is very unlikely to reach $s_B$ so a causal model would not learn to distinguish $s_B$ from states which do not lead to reward. This is similar to the example with the rooster from the Introduction: in case if the agent is located in an area with roosters, it will naturally rely on the rooster to "predict" the sunrise. However, a sequence of hard-to-discover "deliberate" actions (such as moving to a far-away area) will force the model to rely on a more robust feature, such as a measurement of the relative position of the Sun with respect to Earth, even if this measurement is noisy.

In the next subsection we formulate the {\em minimax} causal graph. That definition would favor relying on $f_1$ instead of $f_2$, because there exists a policy (always choosing $a_1$) which gives a very high loss for relying on $f_2$. We say that using $f_2$ to predict the reward is a spurious correlation, since this situation is identical to the example with the rooster from the introduction.

Now, imagine that noise in feature $f_1$ is zero, which would happen if $p=0$. In that case, for any policy that visits $s_B$ at least once during data collection, the optimal model relies on $f_1$. Intuitively, random policy fails for that MDP with $p>0$ simply because it does not visit one of the states often enough. Therefore, the model has a lower loss if it relies on spurious correlations (feature $f_2$).  A way to overcome that is to develop a policy that visits $s_B$ more often. In the next section, we introduce a way to design such policies, which we would call {\em interventions}.

{\bf Causal model of an environment.} For all the cases, we define the problem in a minimax fashion\footnote{Another approach would be to select top-k best causal models and then define the loss as how well at least one of the can predict the data.}: the graph should not be disproved even by the worst policy $\pi$, at any number of collected episodes $t$. This is similar to the scientific method in the real world: we want to learn causal relationships that are true no matter which experiments or actions we perform.

\begin{definition}[Minimax definition of the causal graph]
For an environment $\mu$, we define the causal graph $G^*$ as the one which gives the smallest possible loss $L$ even on the worst policy:
\begin{equation}
G^*=\arg\min\limits_{G}\max\limits_{\pi}L_{\pi}(G)=\arg\min\limits_G\max_{\pi}\lim\limits_{E\to\infty}L_{\pi,E}(G)\label{eq:problem}
\end{equation}
\label{def:minimax}
\end{definition}

In practice, in the equation above, we use a finite sequence of policies, instead of all policies in $\max_\pi$, and we consider a finite $E<\infty$. This can be seen as a sample estimate of the Equation \ref{eq:problem}.

As noted in the previous subsection, the Definition \ref{def:minimax} would favor relying on a noisy feature $f_1$ in the example from Figure \ref{fig:ex_two_path} achieving a constant non-zero loss in all cases, instead of relying on a more risky feature $f_2$ which would only work in $s_C$ but not in $s_B$. In this way, the minimax definition favours "safe" solutions instead of "risky" ones.

Specifically, after executing a set of policies $\pi_1,...,\pi_s$ and obtaining graphs $G_1,...,G_s$, we would like to compute the policy $\pi_{s+1}$ that would obtain as much information to improve $G_{s+1}$ over previous graphs as possible. Executing the next policy $\pi_{s+1}$ after $\pi_s$ can be seen as doing an intervention $do(I_{s+1})$\footnote{Operator $do(\cdot)$ sets \citep{pearl2018book,halpern2005causes,pearl2018theoretical} the value for some nodes in the model, without changing the values of its parents, but changing the values of downstream nodes.} in the causal model, since the policy sets nodes to specific values. In that sense, we sample novel histories from the interventional distribution $\mathbb P[h|do(I_{s+1})]$.
Instead of specifying the policy $\pi_{s+1}$ directly, we specify the reward for that policy, and we call the algorithm that designs this reward the {\em intervention designer}.

In the next section we show how to design interventions. The complete setup is shown in Figure \ref{fig:causalagent}.

\section{Solution}
\label{sec:solution}
Before, we defined a way to obtain a causal graph of a reinforcement learning environment. To deal with spurious correlations, we design the intervention reward based on the current causal graph. Next, we combine data generated from an agent trained on that reward with previously collected data to learn a better causal graph. In the next section we give concrete methods for intervention reward design. These are algorithms that determine which reward to give the agent to come up with a better causal model.

{\bf Intervention design via edges.} We test if edges in the learned graph $G$ are "real" or caused by spurious correlations.  We test an edge $e=f_i\to f_j$ with a positive coefficient in a linear causal model\footnote{In the non-linear case, the coefficient might depend on the current values of features. In that case, this approach will still work but the step has to be small enough to allow for such linearization.}, by setting $do(f_i=\max f, f_j=\min f)$ (minimal and maximal feature values). To do so, we reward the agent for setting $f_i=\max f$ and for $f_j=\min f$. The total reward is $R=f_i-f_j$ \footnote{In case with a negative coefficient in the model, we need $f_i+f_j$.}. When optimizing for it, the agent does its best in executing an intervention $do(f_i=\max f, f_j=\min f)$. This reward will encourage the agent to find data for which $f_i$ has a high value and $f_j$ has a low value. Given that the causal model initially assumed that $f_i$ has a positive effect on $f_j$, such data may disprove the causal model. To obtain the best causal graph, we select edges which the learning algorithm is uncertain about more often than those which the algorithm is certain about. We measure uncertainty by how much different the model predictions are if trained on different subsets of the history.

{\bf Intervention design via nodes.} We reward the agent for setting a target node $f_i$ to a target value $x$. We also reward for "keeping everything else the same". We achieve the latter either by a) penalizing for the difference $d$ between statistics (the averages and variances) of feature distributions from previous and current policies: $R=-|f_i-x|-d$ or b) keep and down-scale the reward from the previous iteration: $R=-|f_i-x|+\gamma R_{old}$. When optimizing for this reward, the agent does its best in executing an intervention $do(f_i=x)$. We choose nodes in the graph based on the total uncertainty of adjacent edges (like in the previous technique, Edge-based intervention design, the uncertainty is computed via the variance over models trained on different subsets of the data).

{\bf Intervention design via the loss.} We reward the agent for finding policies $\pi$ which give high causal graph loss: $R=\sum_i\left(f_i^t-\tilde{f}_i^t\right)^2$. The reasoning behind this is that we want to find data disproving our model, like in Eq. \ref{eq:problem}. Compared to previous methods, we do not select the node or an edge explicitly. This reward design is similar to the curiosity approaches \citep{pathak2017curiosity}\footnote{The difference is that here, we fix the curiosity reward for many time-steps, even when a better model is available. This can be sometimes theoretically more optimal than updating the model as soon as the new data is available \citep{leike2016thompson}}.

To sum up causal learning, in our approach, we simply regress the current time-step $f_t$ on the previous one $f_{t-1}$: $f_t=Wf_{t-1}+b$ using a linear relationship with $l_1$ regularization. It is trained in a (self-)supervised manner by aggregating data from different policies to approximate a solution to Eq. \ref{eq:problem}. We note that better methods \citep{runge2019inferring} of learning causal graphs would still fail without interventions\footnote{Indeed, they have to rely on the observational data. If the data suggests a relationship between features, they have to use it, otherwise they will give a wrong answer on a case where this relationship is actually causal.}. Other causality learning methods are compatible with our approach, since we only require a graph as an output, and we give observational data as input. Methods which discover the features end-to-end \citep{thomas2018disentangling,Kurutach2018,Ke2019,Francois-Lavet2018,Zhang2019} can be used to discover the nodes in the causal graph.

Now, all the discussed components are combined together into a causal agent, to discover the true causal graph of the environment, see Figure \ref{fig:causalagent}.

\section{The environment}
\label{sec:environment}
We use a simple\footnote{While the environment is simple, a random policy or a policy trained with the reward does not obtain the true causal graph in all settings (see \autoref{subsec:results}).} Grid-World environment. The agent needs to eat food in the environment, or the episode will end. In addition, it collects keys to open chests, with each chest giving a reward. There is a button which turns a light on and off and does not give reward. A successful policy needs to balance between staying alive (collecting food), collecting keys and chests (with keys required before opening a chest) and exploring the effect of the button. Figure \ref{fig:envs_causal} represents the causal model we want to discover. Appendix \ref{app:environment} contains more details about the environment.

We use the following specific environments:
\begin{enumerate}[label=(\Alph*)]
\item 5x5 grid-world with randomly placed items.
\item (Figure \ref{fig:envs_b}) a grid-world with a fixed map, where the agent must collect the key before the food.
\item (Figure \ref{fig:envs_c}) 10x10 grid-world with randomly placed items where the food is close to the key, and the chest is far away.
\end{enumerate}

Since the environment A is random, the random policy is expected to uncover the correct graph, as any spurious correlation will disappear on the new layout.

\begin{figure}[h]
    \centering
    \begin{subfigure}[b]{0.49\textwidth}
         \centering
         \includegraphics[width=\textwidth]{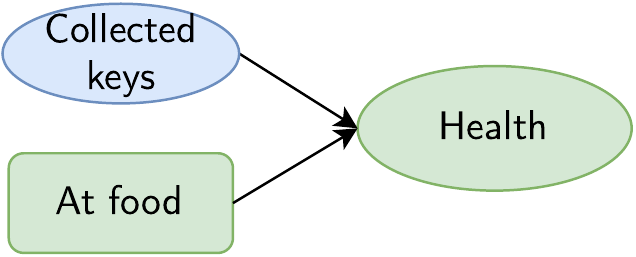}
         \caption{Wrong causal model for Environment C}
         \label{fig:causal_model_correct_incorrect_w}
    \end{subfigure}
    \hfill
    \begin{subfigure}[b]{0.49\textwidth}
         \centering
         \includegraphics[width=\textwidth]{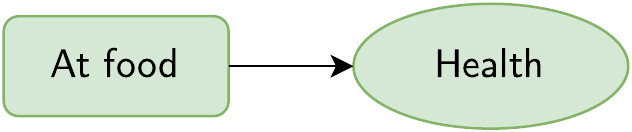}
         \vspace{13pt}
         \caption{Correct causal model for Environment C}
         \label{fig:causal_model_correct_incorrect_c}
    \end{subfigure}
    
    \caption{Two (parts of) models for Environments B, C (from Figure \ref{fig:envs}), in terms of nodes from Figure \ref{fig:envs_causal}. Left (\ref{fig:causal_model_correct_incorrect_w}): the model learned on a random policy or on the policy maximizing for the reward. The collected keys are spuriously correlated with health, because on these policies, it is likely that the agent collects the key before collecting food (collecting {\em food} actually results in an increase in health). Right (\ref{fig:causal_model_correct_incorrect_c}): the model learned using our method. The parent for health is correctly identified as collecting food. 
    }
    \label{fig:causal_model_correct_incorrect}
\end{figure}

In environments B, C it is easy to learn the spurious correlation in which getting a key causes an increase in health (while health is actually determined by the food collected, see Figure \ref{fig:causal_model_correct_incorrect}), because the agent will usually have a key when it collects the food. To uncover the correct model, the agent needs to get rid of the key by opening a chest. After that, it can collect food without having the key, and then the model will rely on the food collected to predict health.

We add noise to each of the environments. With some probability, food is visible at cells not containing food. Thus, we make the environment similar to the MDP from Figure \ref{fig:ex_two_path}. Indeed, there is the easy-to-discover path that leads to the wrong causal model (keys "cause" health to increase), while more exploration (getting rid of the key) yields the correct model (food causes health to increase).

The environment we choose is characteristic of the real world, as it contains spurious correlations that we need to uncover by changing the behavior.

\section{Experiments}

{\bf Hardcoded features.} We augment the feature set with conjunctions of relevant features in order to keep the problem in the linear domain. This allowed us to keep the causal learning simple to focus on interventions. The many techniques \citep{Ke2019,goudet2018learning,chalupka2014visual} for learning non-linear causal graphs that are compatible with our approach, as we only require learning from observational data.

{\bf Baselines and methods.} For all methods we first use a random policy for exploration of the environment. Next, we train a PPO \citep{schulman2017proximal} agent to follow the intervention reward. We compare the three proposed methods for interventions: rewarding the agent proportional to the loss of the model (Loss), disproving edges (Edge), setting nodes to values (Node). We measure if the correct graph was learned using cosine similarity between the node adjacency matrices of the currently learned graph and the ground truth.

\subsection{Results}
\label{subsec:results}
The random and environment-reward policies discover the true causal graph in environment A. Since the environment is randomly generated, there are no spurious correlations. The random and environment-reward policies extremely rarely discover the true causal graph in environments B and C. This is because the food presence feature is noisy and a random policy often collects the food before the keys because they are close together. To predict the health increase, it is best to rely on the spurious feature "food and keys $>0$". This is similar to what happens in the MDP from Figure \ref{fig:ex_two_path} and in the rooster example from the Introduction.

The intervention methods discover the true causal graph in environments B and C more often (results for C in the appendix). This is because they also include data from the intervention policy which collects the food when the agent does not have a key. With data from the intervention policy, it is no longer optimal to rely on the spurious feature: after collecting the key and opening the chest, the key disappears, and the food is collected without the presence of keys.

The Loss intervention method outperforms Node and Edge methods on environment (B). The main problem with the Node and Edge methods is that they have to choose the correct node or edge to intervene on. Once the correct edge or node is chosen, the true graph is learned quickly. In contrast, the Loss method does not have to choose the right thing to intervene on. In simple environments hand-designed methods (edges and nodes) perform reasonably well. However, if we increase the number of features or the complexity of the environment, they stop finding good policies. We did not test the node and edge interventions in environment C extensively due to this reason.

Sampling without replacement when selecting edges gives faster convergence to the true graph versus sampling with replacement. We also found that selecting edges based on uncertainty (by training on different subsets $s=5$) gives better results than selecting random edges. Results for selecting nodes based on uncertainty are similar to selecting randomly. For the nodes method, we found that keeping the new policy close to the old one by using the reward from the environment works. However keeping the feature statistics the same does not work because the agent learns a new way to achieve the same statistics.

\begin{figure}
    \centering
    \includegraphics[width=0.8\textwidth]{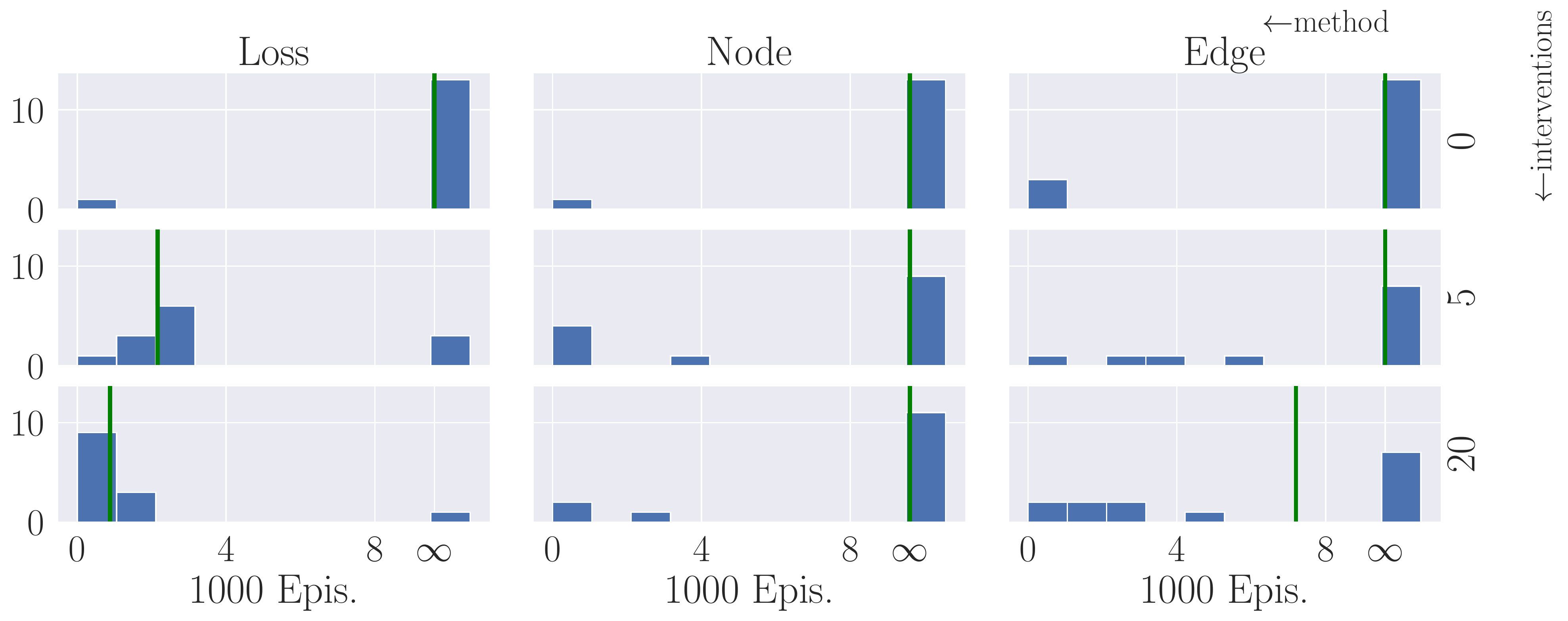}
    \caption{Experimental results on environment B for predicting the true causal graph for the health. Plots are arranged by intervention method (columns: Loss, Node, Edge) and by the number of interventions performed (rows: 0, 5, 20). Each plot shows the histogram for the number of episodes (in 1000s) it takes to find the true causal graph $G^*$. The horizontal axis shows the number of episodes, and the vertical axis represents the number of runs (out of 10) which have converged to the true graph $G^*$. $\infty$ means the graph was not found during training. {\color{green} Green line} represents the median number of episodes it takes to find the correct graph. $0$ interventions corresponds to training with the original environment reward. The random policy is evaluated in a separate experiment with spurious correlations as a result. The overall trend shows that the median number of episodes before uncovering the correct graph decreases for Loss and Edge methods, as the number of interventions increases.}
    \label{fig:env_b}
\end{figure}

\section{Conclusion}
We design a method to learn the causal model of the environment by performing interventions, which helps prevent learning spurious correlations. This shows the potential of RL to improve causal graph learning and compares techniques to accomplish this. We state the problem of learning a causal graph in a RL setting, so other work can build off of ours. 

\section{Future Work}
We plan to combine our graph learning with one of the approaches for learning the features \citep{Kurutach2018,Ke2019,Francois-Lavet2018,Zhang2019} and train the entire network end-to-end. The sparsity loss will help the features be disentangled because it will minimize dependencies between them \citep{thomas2018disentangling}.

To make our method more general, we plan to use one of the advanced non-linear causality learners \citep{Ke2019}. Some of them are differentiable, which would allow to backpropagate from the graph to the features.

Finally, we can utilize the high-level graph as a hierarchical RL controller \citep{nachum2018data}. Specifically, we can run a traversal algorithm on the causal graph to find chains of nodes that lead to high reward. Then, we can reward the agent for activating the nodes in the correct sequence. This might increase the robustness to distributional shift, as we will rely on the correct features for acting.

\section{Acknowledgements}
We thank Ivan Vendrov, Rohin Shah, Jacob Hilton, Honglak Lee, Lars Buesing, Prashan Madumal for helpful discussions and feedback on drafts of this paper. We thank Stuart Russell for pointing us to similar projects. Thanks to Anastasia Koloskova, Olesya Altunina, David Lindner, Todor Markov, Denis Drescher, Dmitry Mironov, Igor Molybog, Valentin Hartmann and Bryce Woodworth for providing feedback on the draft and videos and to El-Mahdi El-Mhamdi for introducing us to the consciousness prior. Thanks to David Krueger for suggesting an improved version of the causal learning objective function.

\bibliography{iclr2020_conference}
\bibliographystyle{iclr2020_conference}

\appendix
\section{Relevant work}
Our method to perform an intervention on an edge is similar to the method used in \citet{Marino2019} to test hypotheses. Compared to that approach, we are interested in the true causal graph of the environment rather than in testing specific hypotheses. Interventions to learn the true graph can be seen in \citep{de2019causal}. Our approach is focused on learning the correct graph rather than acting well. We extend the Action-Influence model \citep{Marino2019,everitt2019understanding} to understand the environment. Compared to \citep{Madumal2019}, we learn the graph rather than design it by hand. The idea to reward the agent for the loss of the causal graph is taken from \citep{pathak2017curiosity}. However, here we are interested in a very low-dimensional causal graph rather than in a black-box model of the environment. Compared to standard model-based techniques \citep{ha2018world}, our approach filters spurious correlations in the model.

\subsection{Relevant work after the first pre-print release}
Compared to \citep{javed2020learning}, we are learning the causal graph of practical RL environments, we define the minimax graph and have a comparison of various intervention techniques.

\section{Hyperparameter selection}
{\bf Resources and parameters.} Parameters were chosen with a hyperparameter search on the task of solving the environment using PPO \citep{schulman2017proximal}. We run the total of 8000 episodes for B and 50000 episodes for C. We vary the number of epochs to train the causal graph in 500-10000, number of interventions 0-50, number of training calls 5-100, intervention method Loss, Edge, maximal number of episodes in the buffer 10-5000, method to select the edge Constant, Weighted and Random. We learn the graph on evaluation data without noise, and update the reward for the trainer.

\section{The environment}
\label{app:environment}
We implement the environment using \href{https://github.com/deepmind/pycolab}{pycolab}. All updates are delayed 1 time-step to give causal information.

An open-source re-implementation can be found in \href{https://github.com/sergeivolodin/causality-disentanglement-rl/tree/master/keychest}{github.com/sergeivolodin/causality-disentanglement-rl}.

\section{Experiments}
\begin{figure}
    \centering
    \includegraphics[width=0.4\textwidth]{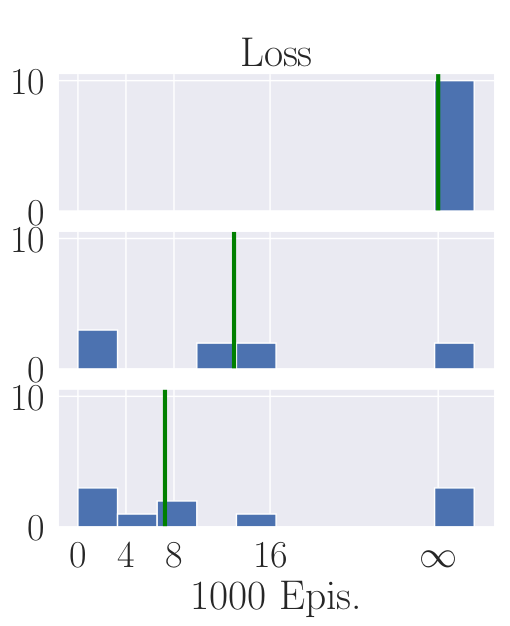}
    \caption{Experimental results on environment C for predicting health, reward, keys and lamp. The description matches that of Figure \ref{fig:env_b}.}
    \label{fig:env_c}
\end{figure}

Figure \ref{fig:env_c} shows the results for environment C. Without interventions, the correct graph is never uncovered. In contrast, with interventions, the correct graph is learned, the more interventions the better.

\section{Realizable case}
\begin{proposition}
\label{prop:noiseless}
For an environment $\mu$ and features $f$, learner $L$ which can always converge to a global minimum for any data distribution, if a true causal graph is $G^*$ (s.t. $L_{\pi}(G^*)=0$ for all $\pi$ and $L>0$ for other $G$), a random policy $\pi_r$ gives the true graph: $\min_G\max_t L_{\pi_r, E}(G)=\min_G\max_{\pi}\max_E L_{\pi, E}(G^*)$
\end{proposition}

Note that in our experiments, we use a linear model which complies with this property. Since a linear model is always convex, its optimization procedure always finds a global minimum. In contrast, we expect that more complex causality learners would benefit significantly from doing directed interventions rather than doing random exploration: Proposition 1 no longer applies to them.

Intuitively, in the realizable case, more data is always better. The proof is based on two ideas: first, for a policy $\pi^*$ giving the true $G^*$, a random policy $\pi_r$ will take same actions as $\pi^*$ with some probability: $\mathbb P[\pi_r=\pi^*]>0$. Thus, data from $\pi^*$ will be in the dataset. Next, since $G^*$ fits any policy, the learner will find a graph $G$ s.t. $L_{\pi_r,E}(G)=0$. By linearity of $L$, loss on $\pi_r$ equals a non-negative combination of losses over policies $\pi_r$ equals to, including one for $\pi^*$. Now, since the non-negative combination is $0$, one particular term $L_{\pi^*,E}(G)=0$ as well, and $G=G^*$. The full proof is below.

\begin{proof}
Consider data (histories) $h$ coming from the agent-environment interaction. We write $h\sim \pi$ meaning that $h$ is a history obtained from interaction between a policy $\pi$ and our fixed environment $\mu$ which we omit in this notation. If $E(h, G)$ corresponds to the error of the model given by a causal graph $G$ on the history $h$, we have by definition $L_{\pi,E}(G)=\frac{1}{E}\sum\limits_{e=1}^E\E(h_e, G)$. If we take $\max_e$, there is a limit $L_{\pi,E}(G)\to \E_{h\sim \pi}\E(h, G)$ as $e\to\infty$: if we add more data to the learner, it is the same as training on "infinite data" in the sense of the distribution.

By definition of $G^*$, we have $\forall G$ $\max_{\pi}\max_E L_{\pi, E}(G)\geq \max_{\pi}\max_E L_{\pi, E}(G^*)$. We define $L_{\pi}(G)=\max_EL_{\pi,E}(G)$ and $L(G)=\max_{\pi}L_{\pi}(G)$, then we have $L(G)\geq L(G^*)$, which means that $G^*$ is the global minimum of $L(\cdot)$ whose numerical value we write as $L_{\min}=L(G^*)$. By the definition of $L(G^*)=\max_{\pi}L_{\pi}(G^*)=L_{\min}\in\mathbb{R}$. Also by the theorem statement, $G^*$ achieves the minimal value of $L$ given any policy: we have $L_{\pi}(G^*)=L_{\min}$ for all $\pi$. This means that $G^*$ is also a global minimum for $L_{\pi}$ for each $\pi$.

Next, consider a random policy $\pi_r$ which is a random variable over the space of all deterministic policies. We write some random outcomes $\omega\in\Omega$ meaning the probability space of all deterministic policies. The interpretation is that by the "inverse" of the Principle of Deferred Decisions (we pretend that we made all the random choices in advance rather than making them as they are required), there are some factors $\omega$ that influence the agent-environment interaction. Once they are chosen, the policy $\pi_r$ is just some deterministic policy for that $\omega$. Then, we can write $L_{\pi_r}(G)\equiv \E_{h\sim \pi_r}E(h, G)=\E_{\omega\in\Omega}\E_{h\sim \pi_r}E(h, G)$. Now, for each outcome $\omega\in\Omega$, $\pi_r$ is a deterministic policy, and then, $\E_{\omega\in\Omega}\E_{h\sim\pi_r}E(h,G)\equiv \E_{\omega\in\Omega}\E_{h\sim\pi_{\omega}}E(h, G)$. Thus, we can write $L_{\pi_r}(G)=\E_{\omega\in\Omega}\E_{h\sim\pi_{\omega}}E(h,G)\equiv \E_{\omega\in\Omega}L_{\pi_{\omega}}(G)$

So now, $L_{\pi_r}(G)=\E_{\omega\in\Omega}L_{\pi_{\omega}}(G)$. Now, since $G^*$ is a global minimum of $L_{\pi}$ for any $\pi$, $L_{\pi_r}(G)\geq \E_{\omega\in\Omega}L_{\pi_{\omega}}(G^*)=L_{\pi_r}(G^*)=L_{\min}$.

Now, consider $L_{\pi_r}(G)$ from another perspective. This corresponds to a model $G$ trained on data from $\pi_r$. If we use the fact that the learner always finds the global minimum, we know that $L_{\pi_r}(G)\leq L_{\pi_r}(G^*)$: we know that on $G^*$ it has a certain loss. Therefore, if it has output $G$, the loss should be less.

Now, we have two inequalities which we merge into one equality: $L_{\pi_r}(G^*)\geq L_{\pi_r}(G)\geq L_{\pi_r}(G^*)$. Thus, we see that $L_{\pi_r}(G)=L_{\pi_r}(G^*)=L_{\min}$.

Finally, the last equality means that $L_{\pi_r}(G)$ attains the minimal possible value of $L$, thus, $G$ also delivers the global minimum of $L(\cdot)$ and $L(G)=L(G^*)$
\end{proof}

\section{MDPs from Figures}
\label{app:pomdp1}

{\bf MDP from Figure \ref{fig:ex_nonrealizable}}
\begin{enumerate}
    \item If $p=0$, we always choose to go to $s_2$ from $s_0$. Therefore, we end up in $s_4$ after that, which corresponds to $o_3$, analogously $p=1$ corresponds to going to $s_1$ and then to $s_3$ which corresponds to $o_2$
    \item If we set $p\in(0,1)$ and write the loss of a linear model with coefficients $w_2$, $w_3$ for the $o_2$ 1-hot component and the $o_3$ 1-hot component respectively, the loss is $L=p\cdot\left[(1-w_2)^2+w_3^2\right]+(1-p)\cdot\left[w_2^2+(1-w_3)^2\right]$. Minimizing this expression gives $w_2=p$, $w_3=1-p$, $L=2p-2p^2$.
    \item If we consider the minimax case, we need to select a graph which results in the smallest loss in the worst case. To do that, we first maximize the previous expression for $L$ over $p$, and then minimize it over $w_2, w_3$, which gives $w_2=w_3=1/2$, $L=1/2$ and does not depend on $p$.
\end{enumerate}

{\bf MDP from Figure \ref{fig:ex_two_path}}
\begin{enumerate}
    \item The linear reward-predicting model must have a bias of $0$ so that it predicts zero rewards in states $s_A,s_1,...,s_n$ (when $n$ is large enough, this becomes significant).
    \item If we take $a_2$ at the beginning, the loss of a linear model predicting the reward $s_C\to s_D$ would be $(1-w_1\xi-w_2\cdot 1)^2$. The expectation of this becomes $(1-p)(1-w_1-w_2)^2+p(1-w_2)^2$. Minimizing this over $w_1,w_2$ gives $w_1=0$ and $w_2=1$: we rely on the feature $f_2$, and the best loss is $L=0$
    \item If we always take $a_1$, the loss for predicting the reward at $s_B\to s_D$ would be $(1-w_1\xi)^2$. Its expectation is $p+(1-p)(1-w_1)^2$, the best loss is $L=p$ and $w_1=1$, $w_2=0$ (if noise parameter $p<1$). We rely on $f_1$, even if it is noisy.
    \item Now, if we take a policy which arrives to $s_B$ with probability $q$, the loss becomes a linear combination of the two losses with coefficients $q$, $1-q$, the weights depend on the coefficients $p, q$. Specifically, consider a policy arriving at $s_B$ with probability $q$. For a random uniform policy, $q=\Theta(1/2^n)$ (up to a constant). Now, the loss becomes $L=q\cdot\left[p+(1-p)(1-w_1)^2\right]+(1-q)\cdot\left[(1-p)(1-w_1-w_2)^2+p(1-w_2)^2\right]$. This function attains a minimum at a point with $w_1=q/(p+q-pq)$. If $p\leq q\alpha$, $w_1=q/(q\alpha+q-q^2\alpha)\geq 1/(1+\alpha)$. For $\alpha=1/100$, we get that $w_1\geq 1-1/100\approx 1$. This shows that for $p\ll q$, $w_1\approx 1$. For a random uniform policy this translates to $p\ll 1/2^n$ which means that $n \ll -\log p$
    \item If we consider the minimax case, we need to select a graph which results in the smallest loss in the worst case, which would mean relying on $f_1$ and obtaining a worst-case loss of $p$ (worst-case $q=0$) instead of relying on $f_2$ and obtaining a worst-case of loss $1$ (worst case $q=1$).
\end{enumerate}

\end{document}

%% file: math_commands.tex
\usepackage{amsmath,amsfonts,bm}

\def\eqref#1{equation~\ref{#1}}
\def\1{\bm{1}}

\DeclareMathAlphabet{\mathsfit}{\encodingdefault}{\sfdefault}{m}{sl}
\SetMathAlphabet{\mathsfit}{bold}{\encodingdefault}{\sfdefault}{bx}{n}

\newcommand{\E}{\mathbb{E}}